\documentclass{article}
\usepackage{ijcai19}

\usepackage{times}
\usepackage{soul}
\usepackage{url}
\usepackage{hyperref}
\usepackage[utf8]{inputenc}
\usepackage[small]{caption}
\usepackage{graphicx}
\usepackage{amsmath}
\usepackage{booktabs}
\usepackage{algorithm}
\usepackage{algorithmic}
\usepackage{subfig}

\usepackage[subtle]{savetrees}

\usepackage{todonotes}

\usepackage{gensymb}
\usepackage{amsfonts}
\newtheorem{problem}{Problem}

\newcommand{\LL}[1]{{\color{red}[LL: #1]}}
\newcommand{\ap}[1]{{\color{blue}[AP: #1]}}
\newcommand{\np}[1]{{\color{violet}[NP: #1]}}
\newcommand{\MW}[1]{{\color{orange}[MW: #1]}}

\newcommand{\extratext}[1]{} 
\newcommand{\dataset}[0]{ \mathcal{D} }
\newcommand{\wv}[0]{ \mathbf{w} }
\newcommand{\numclasses}[0]{ C }
\newcommand{\numweights}[0]{ W }
\title{Statistical Guarantees for the Robustness of Bayesian Neural Networks}

\author{
Luca Cardelli$^1$ \and
Marta Kwiatkowska$^1$\and
Luca Laurenti$^{1}$\thanks{equal contribution}\and
Nicola Paoletti$^2$\and\\
Andrea Patane$^{1}$\footnotemark[1]\And
Matthew Wicker$^{1}$\footnotemark[1]
\affiliations
$^1$ University of Oxford\\
$^2$Royal Holloway University of London
\emails
\{luca.cardelli, marta.kwiatkowska,
luca.laurenti, andrea.patane, matthew.wicker\}@cs.ox.ac.uk,
nicola.paoletti@rhul.ac.uk
}
\begin{document}

\maketitle

\begin{abstract}

We introduce a {probabilistic
robustness} measure for Bayesian Neural Networks (BNNs), defined as the probability that,
given a test point,
there exists a point within a bounded set such that the BNN prediction differs between the two. 
Such a measure can be used, for instance, to quantify the probability of
the existence of adversarial examples. Building on statistical
verification techniques for probabilistic models, we develop a framework that allows us to estimate probabilistic
robustness for a BNN with statistical guarantees, i.e., with \textit{a priori} error and confidence bounds. 
We provide experimental comparison for several approximate BNN inference techniques
on image classification tasks associated to MNIST and a two-class subset of the GTSRB dataset.
Our results enable quantification of uncertainty of BNN predictions in adversarial settings. 
\end{abstract}

\section{Introduction}
\label{submission}

Bayesian Neural Networks (BNNs), i.e.\ neural networks with 
distributions 
over their weights, are gaining momentum for their ability to capture the uncertainty within the learning model, while retaining the main advantages intrinsic to deep neural networks \cite{mackay1992practical,gal2016uncertainty}. 
A wide array of attacks and formal verification techniques have been developed for deterministic (i.e.\ non-Bayesian) neural networks \cite{biggio2018wild}.
However, to date, only methods based on pointwise uncertainty computation have been proposed for BNNs~\cite{feinman2017detecting}.
To the best of our knowledge, there are no methods directed at providing guarantees on BNNs that fully take into account their probabilistic nature. 
This is particularly important in safety-critical applications, where uncertainty estimates can be propagated through the decision pipeline to enable safe decision making \cite{mcallister2017concrete}
.

In this work, we present a statistical framework to evaluate the \textit{probabilistic robustness} of a BNN. 
The method comes with statistical guarantees, i.e., the estimated robustness 
meets \textit{a priori} error and confidence bounds.
In particular, given an input point $x^*\in \mathbb{R}^m$ and a (potentially uncountable) bounded 
set of input points $T\subset \mathbb{R}^m$, 
we aim to compute the probability (induced by the distribution over the BNN weights) that there exists $x \in T$ such that the BNN prediction on $x$ differs from that of $x^*$. 
Note that this is a probabilistic generalisation of the usual statement of (deterministic) robustness to adversarial examples \cite{goodfellow6572explaining}. 

We formulate two variants of probabilistic robustness. 
The first variant describes the probability that the deviation of the network's output (i.e., of the class likelihoods) between $x^*$ and any point in $T$ is bounded. 
This variant accounts for the so-called \textit{model uncertainty} of the BNN, i.e., the uncertainty that derives from partial knowledge about model parameters.
The second variant quantifies the probability that the predicted class label for $x^*$ is invariant for all points in $T$. 
This accounts for both model uncertainty and \textit{data uncertainty}, which is related to intrinsic uncertainty in the labels. 
These properties allow one to estimate, for instance, the probability of the existence of adversarial examples. 

The exact computation of such robustness probabilities is,  unfortunately, infeasible, as the posterior distribution of a BNN is analytically intractable in general. 
Hence, we develop a statistical approach,
based on the observation that each sample taken from the (possibly approximate) posterior weight distribution of the BNN induces a deterministic neural network.
The latter can thus be analysed using existing 
verification techniques for deterministic networks (e.g.\  \cite{huang2017safety,katz2017reluplex,ruan2018reachability}). 
Thus, we can see the robustness of a BNN as a Bernoulli random variable whose mean is the probability that we seek to estimate (see Section~\ref{sec:statistical_method}). 
In order to do so, we develop a sequential scheme based on Jegourel \textit{et al.}~\shortcite{jegourel2018sequential}, a statistical approach for the formal verification of stochastic systems. 
Namely, we iteratively sample the BNN posterior and check the robustness of the resulting deterministic network with respect to the input subset $T$. 
After each iteration, we apply the Massart bounds \cite{massart1990tight} to check if the current sample set satisfies the \textit{a priori} statistical guarantees.
Thus, we reduce the number of samples only to those needed in order to meet the statistical guarantees required.
This is essential for the computational feasibility of the method, as each sample entails solving a computationally expensive verification sub-problem. 
Moreover, our method is generally applicable in that the estimation scheme is independent of the choice of the deterministic verification technique. 
We evaluate our method on fully connected and convolutional neural networks, trained on the MNIST handwritten digits dataset \cite{lecun2010mnist} and a two-class subset of the the German Traffic Sign Recognition Benchmark (GTSRB) \cite{stallkamp2012gtsrb} respectively.
We compare the robustness profiles of three different BNN inference methods (Monte Carlo dropout \cite{gal2016dropout}, variational inference \cite{blundell2015weight}, and Hamiltonian Monte Carlo \cite{neal2011hmc}), showing that our notion of probabilistic robustness results in an effective model selection criterion and provides insights into the benefits of BNN stochasticity in mitigating attacks\footnote{Code is available at\\  \textit{https://github.com/matthewwicker/StatisticalGuarenteesForBNNs}}.

In summary, the paper makes the following main contributions:
\begin{itemize}
\item We define two variants of probabilistic robustness for BNNs, which generalise safety and reachability defined for deterministic networks. 
These can be used to quantify robustness against adversarial examples. 

\item Building on analysis techniques for deterministic neural networks, we design a statistical framework for the estimation of the probabilistic robustness of a BNN, which ensures \textit{a priori} statistical guarantees. 
\item We evaluate our methods on state-of-the-art approximate inference approaches for BNNs on MNIST and GTSRB, for a range of  properties. We
quantify the uncertainty of BNN predictions in adversarial settings.
\end{itemize}


\section{Related Work}
Most existing methods for the analysis and verification of neural networks are designed for deterministic models. 
These can be roughly divided into heuristic search techniques and formal verification techniques.
While the focus of the former is usually on finding an adversarial example  \cite{goodfellow6572explaining,wicker2018feature,wu2018game}, verification techniques strive to formally prove guarantees about the robustness of the network with respect to input perturbations \cite{huang2017safety,katz2017reluplex,ruan2018reachability}. 
Alternatively, statistical techniques posit a specific distribution in the input space in order to derive a quantitative measure of robustness for deterministic networks \cite{webb2018statistical,cohen2019certified}.
However, this approach may not be appropriate for safety-critical applications, because these typically require a worst-case analysis and adversarial examples often occupy a negligibly small portion of the input space. 
Dvijotham \textit{et al.}~\shortcite{dvijotham2018verification} consider a similar problem, i.e., that of verifying (deterministic) deep learning models over probabilistic inputs. Even though they provide stronger probability bounds than the above statistical approaches, their method is not applicable to BNNs. 
Bayesian uncertainty estimation approaches have been investigated as a way to flag possible adversarial examples on deterministic neural networks~\cite{feinman2017detecting}, though recent results suggest that such strategies might be fooled by adversarial attacks designed to generate examples with small uncertainty \cite{grosse2018limitations}. 
\extratext{
In particular, \cite{li2017dropout} evaluate how uncertainty measures estimated by variational dropout are affected by gradient-based attack methods.
\cite{smith2018understanding} analyse how different types of adversarial perturbations affect various measures of uncertainty. 
\cite{feinman2017detecting} rely on density estimation techniques and variational dropout to flag potential out-of-distribution input points.
}
%
However, as these methods build adversarial examples on deterministic network and use uncertainty only at prediction time, their results do not capture the actual probabilistic behaviour of the BNN in adversarial settings. 
In contrast, 
our approach allows for the quantitative analysis of probabilistic robustness of BNNs, yielding probabilistic guarantees for the absence of adversarial examples. 

A Bayesian perspective to adversarial attacks is taken by Rawat \textit{et al.}~\shortcite{rawat2017adversarial}, where experimental evaluation of the relationship between model uncertainty and adversarial examples is given\extratext{computed through an attack guided by a Monte Carlo estimation of the log-likelihood gradient}.
Similarly, Kendall \textit{et al.}~\shortcite{kendall2015bayesian} study the correlation between uncertainty and per-class prediction accuracy in a semantic segmentation problem. 
These approaches are, however, pointwise, in that the uncertainty information is estimated 
for one input at a time. 
Instead, by applying formal verification techniques on the deterministic NNs sampled from the BNN, our method supports worst-case scenario analysis on possibly uncountable regions of the input space. 
This allows us to obtain statistical guarantees on probabilistic robustness in the form of \textit{a priori} error and confidence bounds. 

Cardelli \textit{et al.}~\shortcite{cardelli2018robustness} present a method for computing probabilistic guarantees for Gaussian processes in a Bayesian inference settings, which applies to fully connected BNNs in the limit of infinite width.
However, the method, while exact for Gaussian processes, is only approximate for BNNs and with an error that cannot be computed.

\section{Bayesian Neural Networks}\label{sec:bnn}

In this section we provide background for learning with BNNs, and briefly review the approximate inference methods employed in the remainder of the paper. We use ${f}^\wv(x) = [f_1^\wv(x),\ldots,f_\numclasses^\wv(x)]$ to denote a BNN with $\numclasses$ output units and an unspecified number (and kind) of hidden layers, where $\wv$ is the weight vector random variable.
Given a distribution over $\wv$ and $w \in \mathbb{R}^\numweights$, a weight vector sampled from the distribution of $\wv$, we denote with ${f}^w(x)$ the corresponding deterministic neural network with weights fixed to $w$.
Let $ \dataset =\{(x,c) \, | \, x \in \mathbb{R}^m,\, c \in \{c_1,...,c_\numclasses \} \}$ be the training set. We consider classification with a softmax likelihood model, 
that is, assuming that the likelihood function for observing class $c_h$, for an input $x \in \mathbb{R}^m$ and a given $w \in \mathbb{R}^\numweights$, is given by $\sigma_h \left( f^w(x) \right)= \frac{ e^{f_{h}^w (x)} }{\sum_{j=1}^{\numclasses} e^{f_{j}^w (x)} }$. 
We define $\sigma(f^w(x))=[ \sigma_1\left( f^w(x) \right),..., \sigma_\numclasses \left( f^w(x) \right)]$, the combined vector of class likelihoods, and similarly we denote with $\sigma(f^\wv(x))$ the associated random variable induced by the distribution over $\wv$.
In Bayesian settings, we assume a prior distribution over the weights, i.e.\ $ \wv \sim p(w)$\footnote{Usually depending on hyperparameters,  omitted here for simplicity.}, so that learning for the BNN amounts to computing the posterior distribution over the weights, $p (w \vert \dataset )$, via the application of the Bayes rule. 
Unfortunately, because of the non-linearity generally introduced by the neural network architecture, the computation of the posterior cannot be done analytically \cite{mackay1992practical}.
Hence, various approximation methods have been investigated to perform inference with BNNs in practice.
Among these methods, in this work we consider 
Hamiltonian Monte Carlo \cite{neal2011hmc}, variational inference through Bayes by Backprop  \cite{blundell2015weight}, and Monte Carlo Dropout \cite{gal2016uncertainty}.
We stress, however, that the method we present is independent of the specific inference technique used, as long as this provides a practical way of sampling weights $w$ from the posterior distribution of $\wv$ (or an approximation thereof).

\textbf{Hamiltonian Monte Carlo (HMC)} 
proceeds by defining a Markov chain whose invariant distribution is $p(w \vert \dataset)$, and relies on Hamiltionian dynamics to speed up the exploration of the space.
Differently from the two other methods discussed below, HMC does not make any assumptions on the form of the posterior distribution, and is asymptotically correct.
The result of HMC is a set of samples $w_i$ that approximates  $p(w \vert \dataset)$.

\textbf{Variational Inference (VI)} 
proceeds by finding a suitable approximating distribution $q(w) \approx p(w \vert \dataset)$ in a trade-off between approximation accuracy and scalability.
The core idea is that $q(w)$ depends on some hyper-parameters that are then
iteratively optimized by minimizing a divergence measure between $q (w)$ and $ p (w | \dataset)$. 
Samples can then be efficiently extracted from $q(w)$.

\textbf{Monte Carlo Dropout (MCD)} 
is an approximate variational inference method based on dropout \cite{gal2016dropout}.  
The approximating distribution $q(w)$ takes the form of the product between Bernoulli random variables and the corresponding weights. 
Hence, sampling from $q(w)$ reduces to sampling Bernoulli variables, and is thus very efficient.

 
%
%

\section{Problem Formulation}
A BNN defines  a stochastic process {whose randomness comes from the distribution over the weights of the neural network}. 
Thus, the probabilistic nature of a BNN should be taken into account when studying its robustness.

For this purpose, we formulate two problems, instances of probabilistic reachability and probabilistic safety, properties that are widely employed for the analysis of stochastic processes \cite{abate2008probabilistic,bortolussi2016approximation}. 
At the same time, these problems constitute a probabilistic generalization of the reachability \cite{ruan2018reachability} and safety specifications \cite{huang2017safety} typical of deterministic neural networks. 
In particular, in Problem \ref{ProbFormRegr} we consider reachability of the value of the softmax regression, while Problem \ref{ProbFormCLass} is concerned with perturbations that affect the classification outcome. 



\begin{problem}
\label{ProbFormRegr}
Consider a neural network $f^\wv$ 
with training dataset $\mathcal{D}$. Let $x^*$ be a  test point  and $T\subseteq \mathbb{R}^m$ a bounded set. For a given $\delta\geq 0$, compute the probability
\begin{align*}
&p_1 = \ P(\phi_1(f^\wv) \, | \, \mathcal{D} ), \text{ where}\\
&\phi_1(f^\wv)= \ \exists \, x \, \in T \; s.t. \; |\,\sigma(f^{\wv}(x^*))\,-\, \sigma(f^{\wv}(x))\,|_p \,>\, \delta, 
\end{align*}
and $|\cdot|_p$ is a given seminorm.
For $0\leq \eta \leq 1$, we say that $f^\wv$ is \emph{robust} with probability at least $1-\eta$ in $x^*$  with respect to set $T$ and perturbation $\delta$ iff $ p_1 \leq \eta$.
\end{problem}

For a set $T$ and a test point $x^*$, Problem \ref{ProbFormRegr} seeks to compute the probability that there exists $x \in T$ such that the output of the softmax layer for $x$ deviates more than a given threshold $\delta$ from the output for $x^*$. Note that $x^*$ is not necessarily an element of $T$. 
If $T$ is a bounded region 
around $x^*$, Problem~\ref{ProbFormRegr} corresponds to computing the robustness of $f^{\wv}$ with respect to local perturbations. Note that we only require that $T$ is bounded, and so $T$ could also be defined, for instance, as a set of vectors derived from a given attack. 
The probability value in Problem \ref{ProbFormRegr} is relative to the output of the softmax layer, i.e., to the vector of class likelihoods, and not to the classification outcome, which is instead considered in Problem \ref{ProbFormCLass}. 
%
%
%
In fact, probabilistic models for classification further account for the uncertainty in the class prediction step by placing a Multinoulli distribution on top of the softmax output \cite{gal2016uncertainty}.
Specifically, the class of an input $x^*$ is assigned by the stochastic process $\mathbf{m}(x^*) $ with values in $\{1,...,\numclasses\}$, 
 where the probability that $\mathbf{m}(x^*)=h,$ $h \in \{1,...,\numclasses \}$, is given by   
\begin{equation*}\label{eq:multinoulli}
    P(\mathbf{m}(x^*)=h )=\int z P(\sigma_h (f^\wv(x^*))=z\, | \, \mathcal{D} )dz.
\end{equation*} 
Taking the classification aspect and Multinoulli distribution into account poses the following problem.

\begin{problem}
\label{ProbFormCLass}
Consider a neural network $f^\wv$ 
with training dataset $\mathcal{D}$. Let $x^*$ be a  test point  and $T\subseteq \mathbb{R}^m$ a bounded set. We compute the probability
\begin{align*}
&p_2 = \ P(\phi_2(f^\wv) \, | \, \mathcal{D} ), \text{ where}\\
&\phi_2(f^\wv)= \exists \, x \, \in T \;   s.t. \; \mathbf{m}(x^*)\neq \mathbf{m}(x).
\end{align*}
For $0\leq \eta \leq 1$, we say that $f^\wv$ is \emph{safe} with probability at least $1-\eta$ in $x^*$  with respect to set $T$ iff $ p_2 \leq \eta$.
\end{problem}


An important consequence of Problem \ref{ProbFormCLass} is that, for regions of the input space where the model is unsure which class to assign (i.e., where all classes have similar likelihoods), it is  likely that the Multinoulli samples of Eqn  \eqref{eq:multinoulli} induce a classification different from that of $x^*$, thus leading to a low probability of being safe. In contrast, Problem \ref{ProbFormRegr} does not capture this aspect as it considers relative variations of the class likelihoods. 


Note that the only source of uncertainty contributing to the stochasticity of Problem \ref{ProbFormRegr} comes from the distribution of the weights of the BNN, i.e.\ from $p(\wv | \dataset)$.
This is the so-called \textit{model uncertainty}, i.e.\ the uncertainty that accounts for our partial knowledge about the model parameters \cite{gal2016uncertainty}. 
On the other hand, Problem \ref{ProbFormCLass} accounts both for \textit{model uncertainty} and \textit{data uncertainty}, i.e.\ the noise of the modelled process.
We stress that both robustness measures introduced in Problem 1 and 2 do not consider any specific decision making procedure, and are as such independent and prior to the particular decision making techniques placed on top of the Bayesian model.

 Unfortunately, for BNNs, the distribution of $f^\wv$ is intractable. Hence, to solve Problem~\ref{ProbFormRegr} and~\ref{ProbFormCLass} approximation techniques are required. 
In what follows, we illustrate a statistically sound method for this purpose.
\extratext{Exact computation of Problem \ref{ProbFormRegr} and \ref{ProbFormCLass} boils down to computing the \emph{excursion probability} of a stochastic process, that is, the distribution of the supremum of a stochastic process in a given set~\cite{cardelli2018robustness}. Unfortunately, excursion probabilities cannot  be computed for a general stochastic process and even probabilistic bounds exist only for a few limited class of processes \cite{adler2009random}. Moreover, for BNNs, the distribution of $f^\wv$ is intractable. Hence, in order to solve Problem~\ref{ProbFormRegr} and~\ref{ProbFormCLass}, approximation techniques are needed. 
In what follows, we illustrate a  statistically sound method for this purpose.}

\section{Estimation of BNN Robustness Probability}\label{sec:statistical_method}

We present a solution method to Problems~\ref{ProbFormRegr} and~\ref{ProbFormCLass}, which builds on a sequential scheme to estimate the probability of properties $\phi_1$ and $\phi_2$, (roblems 1 and 2, respectively. This solution comes with \textit{statistical guarantees}, in that it ensures arbitrarily small estimation error with arbitrarily large confidence.  

Our method is based on the observation that a sample of the BNN weights induces a deterministic NN.
Hence, we can decide the satisfaction of $\phi_1$ and $\phi_2$ for each sample using existing formal verification techniques for deterministic networks.
More precisely, given a BNN $f^{\bf w}$, we verify
$\phi_1$ and $\phi_2$ over deterministic NNs $f^{w}$, where $w$ is a weight vector sampled from $p(w | \dataset)$ (or its approximation $q(w)$). 
For $\phi_2$, along with $w$, we also need to sample the class $\mathbf{m}(x^*)$ from the Multinoulli distribution of
Eqn~\eqref{eq:multinoulli}.
Details about the deterministic verification methods used here are given in Section~\ref{sec:det_verification}.

Therefore, for $j=1,2$, we can see the satisfaction of $\phi_j(f^{\bf w})$ as a \emph{Bernoulli random variable} $Z_j$, which we can effectively sample as described above, i.e., by sampling the BNN weights and formally verifying the resulting deterministic network. 
Then, solving Problem $j$ amounts to computing the expected value of $Z_j$. That is, we evaluate the probability $p_j$ that $\phi_j$ is true w.r.t.\ $f^{\bf w}$. For this purpose, we derive an estimator $\hat{p}_{j}$ for $p_j$ such that: 
\begin{equation}\label{eq:prob_estimator}
\hat{p}_{j} = \frac{1}{n}\sum_{i=1}^n \phi_j(f^{w_i}) \approx \mathrm{E}[Z_j] = p_j, 
\end{equation}
where $\{f^{w_i}\}_{i=1,\ldots,n}$ is the collection of sampled deterministic networks. 

We want $\hat{p}_{j}$ to satisfy \textit{a priori} statistical guarantees. Namely, for arbitrary absolute error bound $0 < \theta < 1$ 
and confidence $0< \gamma \leq 1$ (i.e., the probability of producing a false estimate), the following must hold:
\begin{equation}\label{eq:est_constraints}
P(|\hat{p}_{j}-p_j|>\theta)\leq \gamma.
\end{equation}

Chernoff bounds~\cite{chernoff1952measure} 
are a popular technique 
to determine the sample size $n$ required to satisfy~\eqref{eq:est_constraints} for a given choice of $\theta$ and $\gamma$. Specifically, the estimate $\hat{p}_{j}$ after $n$ samples satisfies~\eqref{eq:est_constraints} if 
\begin{equation}\label{eq:chernoff}
n > \frac{1}{2\theta^2}\log\left(\frac{2}{\gamma}\right).
\end{equation}
These bounds are, however, often overly conservative, leading to unnecessarily large sample size. Tighter bounds were formulated by Massart~\shortcite{massart1990tight}, where the sample size depends on the unknown probability to estimate, $p_j$. In particular, Massart bounds require only a small fraction of samples when $p_j$ is close to $0$ or $1$, 
but are not directly applicable for their dependence on the unknown $p_j$ value. Jegourel \textit{et al.}~\shortcite{jegourel2018sequential} solve this issue by extending Massart bounds to work with confidence intervals for $p_j$ instead of $p_j$ itself. 
For arbitrary $0 < \alpha < \gamma$, let $I_{p_j}=[a,b]$ denote the $1-\alpha$ confidence interval for $p_j$ obtained after $n$ samples. 
Then, \eqref{eq:est_constraints} holds if $n$ satisfies
\begin{equation}\label{eq:massart}
\footnotesize
n > \frac{2}{9\theta^2}\log\left(\frac{2}{\gamma-\alpha}\right)\cdot 
\begin{cases}
(3b + \theta)(3(1-b) - \theta) & \text{if } b < 1/2\\
(3(1-a) + \theta)(3a + \theta) & \text{if } a > 1/2\\
(3/2 + \theta)^2 & \text{otherwise}
\end{cases}.
\end{equation}
We employ 
a sequential probability estimation scheme to solve Problem \ref{ProbFormRegr} and \ref{ProbFormCLass}, which utilizes the above bounds to determine, after each sample, if the current estimate provides the required guarantees (given by parameters $\theta$, $\gamma$, and $\alpha$). By applying a sequential scheme, we crucially avoid unnecessary sampling because the analysis terminates as soon as the statistical guarantees are met. 
This considerably improves the efficiency of our method, given that drawing of each Bernoulli sample entails solving a potentially computationally expensive NN verification problem. 

The estimation scheme is outlined in Algorithm~\ref{alg:problem2} and works as follows. At the $n$-th iteration, we sample a weight vector $w$ from the posterior and, only for Problem~\ref{ProbFormCLass}, the class of the test input $\mathbf{m}(x^*)$ (lines 4-5). The $n$-th Bernoulli sample (variable $\mathsf{SAT}$ in line 6) is obtained by applying a suitable deterministic verification method (see Section~\ref{sec:det_verification}) on $f^w$. 
After updating the number of successes $k$ and trials $n$ (line 7), we use these to update the estimator $\hat{p}$ (line 8) and compute a $1-\alpha$ confidence interval $I_p$ for the robustness probability (line 9).
We use $I_p$ to derive the sample size $n^M$ as per the Massart bounds of Equation~\eqref{eq:massart} (line 10), and update the number of required samples to $n_{\max} = \lceil\min(n^M,n^C)\rceil$ (line 11), where $n^C$ is the sample size computed as per~\eqref{eq:chernoff}, in line 1 of the algorithm. In other words, we select the best between Chernoff and Massart bounds, as Massart bounds are tighter than Chernoff bounds when $\hat{p}$ is close to 0 or 1, but Chernoff bounds perform better when $\hat{p}$ is close to 0.5. 
If $n \geq n_{\max}$, we return $\hat{p}$, which is guaranteed to satisfy~\eqref{eq:est_constraints}. Otherwise, we iterate over an additional sample. 
Concrete values of the Chernoff and Massart bounds for our experiments are reported in Appendix~\ref{app:bounds}. 

\begin{algorithm}
\caption{BNN robustness estimation}\label{alg:problem2}
\textbf{Input:} $x^*$ -- test point, $T$ -- search space for adversarial inputs, $f$ -- network architecture, $p(w | \dataset)$ -- posterior on weights, $\phi$ -- property ($\phi_1$ or $\phi_2$), $\theta, \gamma, \alpha$ -- Massart bound parameters\\
\textbf{Output:} $\hat{p}$ -- robustness probability estimator  satisfying~\eqref{eq:est_constraints} 
\begin{algorithmic}[1]
\STATE $n^C \gets$ number of samples by Chernoff bounds~\eqref{eq:chernoff}
\STATE $n_{\max}\gets \lceil n^C \rceil$; $n, k \gets$ 0, 0
\WHILE{$ n < n_{\max}$}
\STATE $w \gets$ sample from $p(w | \dataset)$
\STATE $\mathbf{m}(x^*) \gets$ sample class from \eqref{eq:multinoulli} (only for Problem~\ref{ProbFormCLass})
\STATE $\mathsf{SAT}\gets$ verify $\phi$ over $f^w$ as per \S \ref{sec:det_verification}
\STATE $k \gets k + \mathsf{SAT}$; $n \gets n+1$
\STATE $\hat{p} \gets$ $k/n$
\STATE $I_{p} \gets$ \textsc{ConfidenceInterval}($1-\alpha$, $k$, $n$)
\STATE $n^M \gets$  samples num.\ by Massart bounds~\eqref{eq:massart} using~$I_{p}$
\STATE $n_{\max}\gets \lceil \min(n^M, n^C)\rceil$
\ENDWHILE
\STATE return $\hat{p}$
\end{algorithmic}
\end{algorithm}



\subsection{Verification of Deterministic NNs}\label{sec:det_verification}
Our estimation algorithm is independent of the choice of $T$ and the deterministic verification method used. Below we describe two
configurations that are relevant for robustness analysis of real-world BNNs.

\paragraph{Robustness to Bounded Perturbations.}
In this configuration, $T$ is defined as a ball around the input test point $x^*$. We check whether there exists $x\in T$ such that $\phi_j$ holds for the deterministic NN, $f^w$.
The verification is parametrised by $\epsilon$, the radius of $T$.
We apply the reachability method of Ruan \textit{et al.}~\shortcite{ruan2018reachability} that computes a safe enclosure of the NN output over $T$, along with two points in $T$ that respectively minimize and maximize the output of $f^w$ over $T$.  
Note that the worst-case input over all points in $T$ is one of these two extremum points: for Problem~\ref{ProbFormRegr}, it is the point with the largest likelihood discrepancy from $x^*$; for Problem~\ref{ProbFormCLass}, it is the point that minimizes the likelihood of the nominal class $\mathbf{m}(x^*)$. 
Thus, we can proceed to verify $\phi_j$ by simply checking the property for the corresponding worst-case input. 

\paragraph{Robustness to adversarial attacks.} 
We seek to assess the vulnerability of the network against known attack vectors.
We use the white-box methods by Goodfellow \textit{et al.}~\shortcite{goodfellow6572explaining} and Madry \textit{et al.}~\shortcite{madry2017pgd}.
These work by building the network gradient and transversing the input space toward regions of reduced classification confidence.
Both are parameterised by an attack strength parameter $\epsilon$, used to scale the gradient magnitude. 
Note that the attack is applied to each realisation of the BNN, and as such each attack vector is optimized specifically for $f^w$, for each $w\sim p(w | \dataset)$.

\section{Results}\label{sec:results}

%

We evaluate our method on different BNN architectures trained with different probabilistic inference techniques (HMC, VI, MCD -- see Section~\ref{sec:bnn}). In Section~\ref{sec:verification_res}, we analyse robustness to bounded perturbations for a two class subset of the MNIST dataset.  In Section~\ref{sec:empirical_res}, we report results for adversarial attacks on the full MNIST dataset and a subset of GTSRB. 

\begin{figure*}[h]
\centering
\includegraphics[width=.8\textwidth]{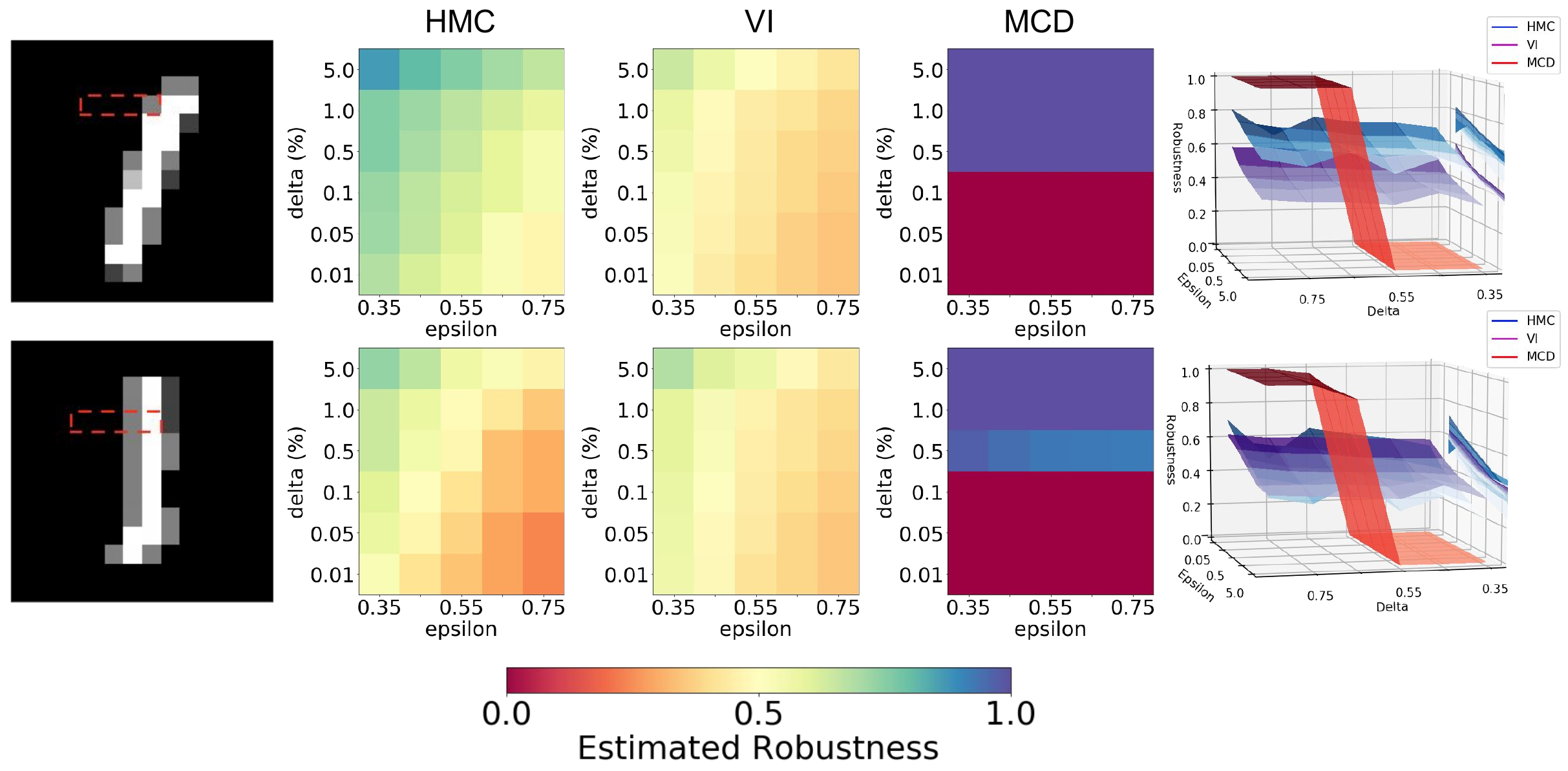}
\caption{ 
On the left are two images from the MNIST dataset with the features to be tested outlined in red. The three central columns contain the heatmaps showing the robustness probabilities (as per Problem 1) for different values of $\epsilon$ ($x$ axis) and $\delta$ ($y$ axis) for BNNs trained with HMC, VI and MCD. On the right are 3D surface plots of the heatmaps with the position of the surfaces projected onto the $yz$-axis so that the heatmaps are easily comparable.
}\label{fig:verification-heatmaps}
\end{figure*}

\subsection{Experimental Settings}
We focus our experiments on BNNs with ReLU activation functions and independent Gaussian priors over the weights.

\begin{figure*}[h]
\centering
\includegraphics[width=.85\textwidth]{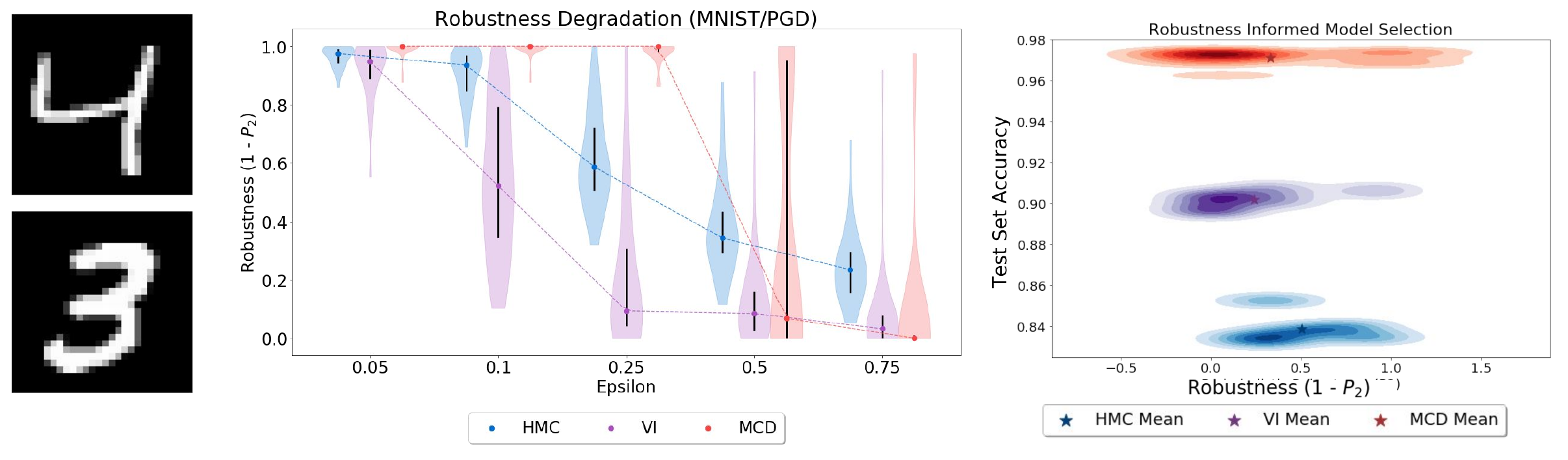}
\caption{On the left we show a two samples from the MNIST training data set. In the center, each violin plot comes from the estimated robustness of 50 different input samples from the network (the same for each violin). On the right, we plot robustness to inform model selection and can observe that MCD robustness peaks are centred at 0 and 1, whereas VI and HMC are more centred.}\label{fig:attacks}
\end{figure*}

In Section~\ref{sec:verification_res} we train a fully conneted network (FCN) with 512 hidden nodes on a two-class subset of MNIST (classes one and seven) and then in Section~\ref{sec:empirical_res} we use the entire dataset.
%
%
%
Also in Section~\ref{sec:empirical_res}, we analyse a two-layer convolutional BNN on a two-class subproblem of GTSRB (examples of the two classes can be seen in the left column of Figure \ref{fig:attacks}). Namely, the convolutional layer contains 25 filters (kernel size: 3 by 3) followed by a fully-connected layer of 256 hidden nodes. Overall, this BNN is characterised by over four million trainable weights. Unfortunately, applying HMC to larger networks is challenging \cite{neal2011hmc}.

For the statistical estimation of robustness probabilities, we used the following Massart bounds parameters: $\theta = 0.075$, $\gamma = 0.075$ and $\alpha = 0.05$. 
Details on training procedures and hyperparameters are included in the Supplementary Materials.

%
%
%
\subsection{Robustness to Bounded Perturbations}
\label{sec:verification_res}
%
%
%


\begin{figure*}[h]
\centering
\includegraphics[width=.85\textwidth]{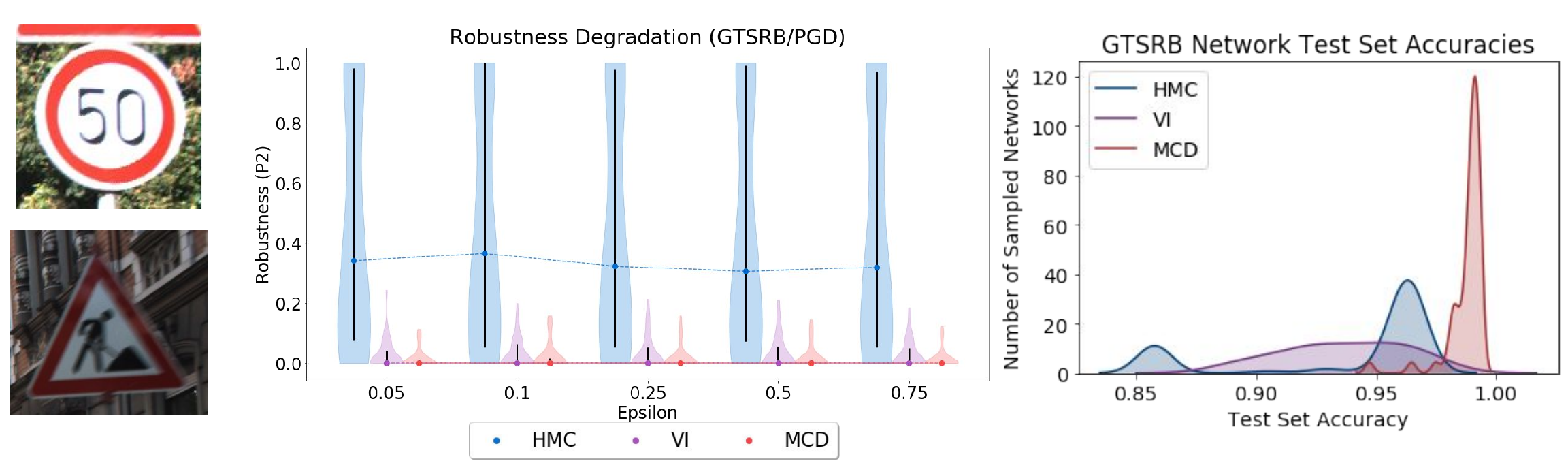}
\caption{In the same spirit as Figure \ref{fig:attacks}, we explore the effect of attack strength on the probabilistic robustness of BNNs trained on GTSRB. On the left we show examples from the two classes tested. 
Unsurprisingly, convolutional neural networks are less robust to attacks, as previously observed in \protect\cite{carlini2017cwattack}. On the right, we plot the accuracy of samples from network posteriors. This information may be used to reason about model selection. }\label{fig:GTSRB-Attacks}
\end{figure*}


Figure~\ref{fig:verification-heatmaps} depicts the results obtained for Problem \ref{ProbFormRegr} on two input images randomly selected from the subset of the MNIST dataset, when relying on Ruan \textit{et al.}~\shortcite{ruan2018reachability} for the deterministic verification sub-routine. 
The input region $T$ is defined as a hyper-rectangle with edge length $\epsilon$ around the reference test image $x^*$. 
In view of scalability limitations of the underlying deterministic method, we restrict the evaluation to the feature highlighted in red in the left column of Figure \ref{fig:verification-heatmaps}. 
We investigate how the robustness probability is affected by variations of $\epsilon$ and $\delta$. 

First of all, note that the estimated robustness decreases as $\epsilon$ increases and/or as $\delta$ decreases, as these respectively imply larger regions $T$ and/or tighter constraints on the BNN output values. 
Interestingly, even in this simple network, the robustness profile strongly depends on the approximation method used for computation of the BNN posterior. 
In fact, for HMC and VI (respectively second and third column of the figure) we observe smooth changes in the robustness probability w.r.t.\ $\epsilon$ and $\delta$, where these changes are quantitatively more prominent for HMC than for VI. 
As with HMC no assumption is made on the form of the posterior distribution, this quantitative robustness difference might suggest that the normality assumption made by VI during training is not sufficient 
in adversarial settings -- i.e.\ as the model is pushed toward corner-case scenarios. 
In turn, this could make the BNN vulnerable to low-variance adversarial examples \cite{grosse2018limitations}.
%
%
On the other hand, MCD (fourth column of Figure \ref{fig:verification-heatmaps}) is characterised by an almost deterministic behaviour with respect to Problem \ref{ProbFormRegr}, with estimated robustness probabilities sharply moving from 1 to 0. This is especially visible when compared with the other two inference methods (fifth column of Figure \ref{fig:verification-heatmaps}). 
As the accuracy scores obtained by the three methods are similar, our results seem to suggest that the BNNs trained by MCD behave almost \textit{deterministically} with respect to \textit{probabilistic} robustness. 
Again, this may be due to the fact that, in adversarial settings, the MCD approximation could lead to an underestimation of model uncertainty. Underestimation of the uncertainty for MCD has also been observed in non-adversarial setting by Myshkov \textit{et al.}~\shortcite{myshkov2016posterior}.


\subsection{Robustness to Adversarial Attacks}\label{sec:empirical_res}

We analyse the resilience of Bayesian CNNs against adversarial attacks.
As robustness of convolutional neural networks is generally defined in terms of misclassification, we provide results for Problem \ref{ProbFormCLass}.

\subsubsection{MNIST}
In Figure \ref{fig:attacks} (central column) we inspect how the BNN behaves under gradient-based attacks with respect to varying attack strength parameterized by $\epsilon$.
The empirical distribution of robustness values shown in the violin plots was estimated by performing statistical verification on 50 images randomly selected from the MNIST test dataset (the empirical average and standard deviation are respectively depicted by a dot and a line centred around it). 
Results serve to butress the observations made in the previous section.
Again, we see that robustness values for VI and, especially, MCD are more stretched toward 0-1 values, compared to those obtained for HMC.
Interestingly, for strong attacks (i.e.\ high values of $\epsilon$), this leads to a relatively higher robustness for MCMC, while small strength attacks consistently fail for MCD.
Thus, it appears that
MCD could be a valuable alternative to MCMC for relatively weak attacks, but may quickly lose its advantage for strong attacks. 
Notice that HMC is the only method consistently showing  
high probability density around the mean value, suggesting that the uncertainty estimation obtained from the posterior could be used in this cases to flag potential adversarial inputs. 

Figure \ref{fig:attacks} right column shows how knowledge of network robustness can be used to select models according to the desired levels of prediction robustness and accuracy. 
For example, if one is not concerned with corner-case scenarios, 
then standard accuracy maximization would have us choose MCD; however, in a case where we are making risk-sensitive decisions, then we might prefer a model that captures a more complete approximation of uncertainty under an adversarial framework - hence trading accuracy for robustness. 

Finally, notice that the computational time required for the statistical verification of an image averages 120 seconds in the experiments here provided.

\subsubsection{GTSRB}

For GTSRB, it is unsurprising that the CNN models are much less robust than the FCNs used for MNIST. It has been shown previously that for networks trained on ImageNet, the $\epsilon$ needed to reduce the test set accuracy to 0\% was 0.004 \cite{carlini2017cwattack}. Despite this, it appears that the architecture trained with HMC maintains some robustness to such perturbations. While it displays the best probabilistic safety across all tests, it has less accuracy in some cases than MCD and VI. This clearly brings up the same trade-offs that were discussed for MNIST.

\section{Conclusion}
We introduced 
probabilistic robustness for BNNs that takes into account both model and data uncertainty, and can be used to capture, among other properties, the probability of
the existence of 
adversarial examples.
We developed a sequential scheme to estimate such a probability with \textit{a priori} statistical guarantees on the estimation error and confidence and evaluated it on fully connected and convolutional networks.
Our methods allows one to quantify the tradeoff between accuracy and robustness for different inference procedures for BNNs. 
We believe our work represents an important step towards the application of neural networks in safety-critical applications. 

\section*{Acknowledgements}\small
This work has been partially supported by a Royal Society Professorship, by the EU's Horizon 2020 program under
the Marie Sk\l{}odowska-Curie grant No 722022 and by the EPSRC Programme Grant on Mobile Autonomy
(EP/M019918/1).



\bibliographystyle{named}
\bibliography{ijcai19}

\newpage
\appendix
\section{Supplementary Material}
In what follows we report the supplementary material of the paper.
We first give specific details on the parameter settings for the inference procedures used for BNN posterior approximation.
We then provide additional results using heuristic adversarial attacks.

\subsection{Experimental Settings}
We report details on the training procedure for the three inference methods analysed in the main text.
\paragraph{HMC.} 
We utilised the implementation of HMC provided in the Edward Python package \cite{tran2016edward}.
We used an  update step size of 0.01 and the numerical integrator was given 5 update steps per sample.

The Gaussian priors on the convolutional filters were set to have mean 1 and variance 0.01 and the Gaussian priors of the fully connected layer were centred at 0 with variance 1.

The fully connected network for MNIST obtained roughly 87\% accuracy on the test set. 
The CNN trained on the GTSRB dataset comprises about 4.3 million parameters, and reaches around 92\% accuracy on the test set.

\paragraph{VI.}
For our implementation of this method we again relied on the Edward Python package \cite{tran2016edward}, using minimisation of the KL divergence.
For the network on GTSRB we train using a batch size of 128, and we use the Adam optimizer with a 0.001 learning rate over 15000 iterations.
These parameters are identical to the training parameters used for MNIST with the exception of training for 5000 iterations with a higher learning rate of 0.01.  
The priors were set accordingly to the ones used for HMC.
\paragraph{MCD.}
In order to choose a dropout value, we grid search parameter of the Bernoulli that governs the dropout method and select the value that gives highest test set accuracy. 
This resulted in the $0.5$ drop-out rate in the BNN used for MNIST, $0.25$ and $0.5$ respectively for the two layers that make up the CNN used for GTSRB.

\subsection{Number of Posterior Samples for Robustness Probability Estimation}\label{app:bounds}
\begin{figure}
    \centering
    \includegraphics[width=.8\columnwidth]{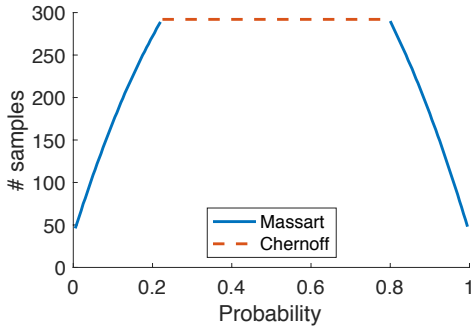}
    \caption{Number of posterior samples (y axis) required in our estimation scheme as a function of the robustness probability. We consider the best of Massart (blue) and Chernoff (orange) bounds. Parameters are $\theta = 0.075$, $\gamma = 0.075$ and $\alpha = 0.05$.}
    \label{fig:bounds}
\end{figure}

Figure~\ref{fig:bounds} shows the number of BNN posterior samples required such that the estimated robustness probability satisfies the following statistical guarantees: error bound $\theta = 0.075$ and confidence $\gamma = 0.075$. These are the parameters used in all our experiments. 

Recall from Section 5 that in our estimation scheme the number of samples $n$ depends on the true probability value $p$,  
and that we employ a more conservative version of the Massart bounds that use the $(1-\alpha)$ confidence interval for $p$, $I_p = [a,b]$, instead of $p$ itself, which is obviously unknown. The x axis of the plot indeed does not represent the true value of $p$, but it represents one of the two extrema of $I_p$ ($a$ when $a>0.5$, $b$ when $b<0.5$, see Equation~\ref{eq:massart}). Indeed we observe that, for probabilities approximately between 0.23 and 0.79, the probability-independent Chernoff bounds are tighter than the (conservative) Massart bounds. Nevertheless, for probability values close to 0 or 1, the Massart bounds are considerably less conservative than the Chernoff bounds, which require 292 samples. For instance, with the above parameters and $\alpha=0.05$, when $b<=0.1$, we need at most 171 samples; when $a>=0.9$, at most 181 samples.

\subsection{Quantitative Difference between Problem 1 \& 2}

\begin{figure}[h]
\centering
\includegraphics[width=0.95\columnwidth]{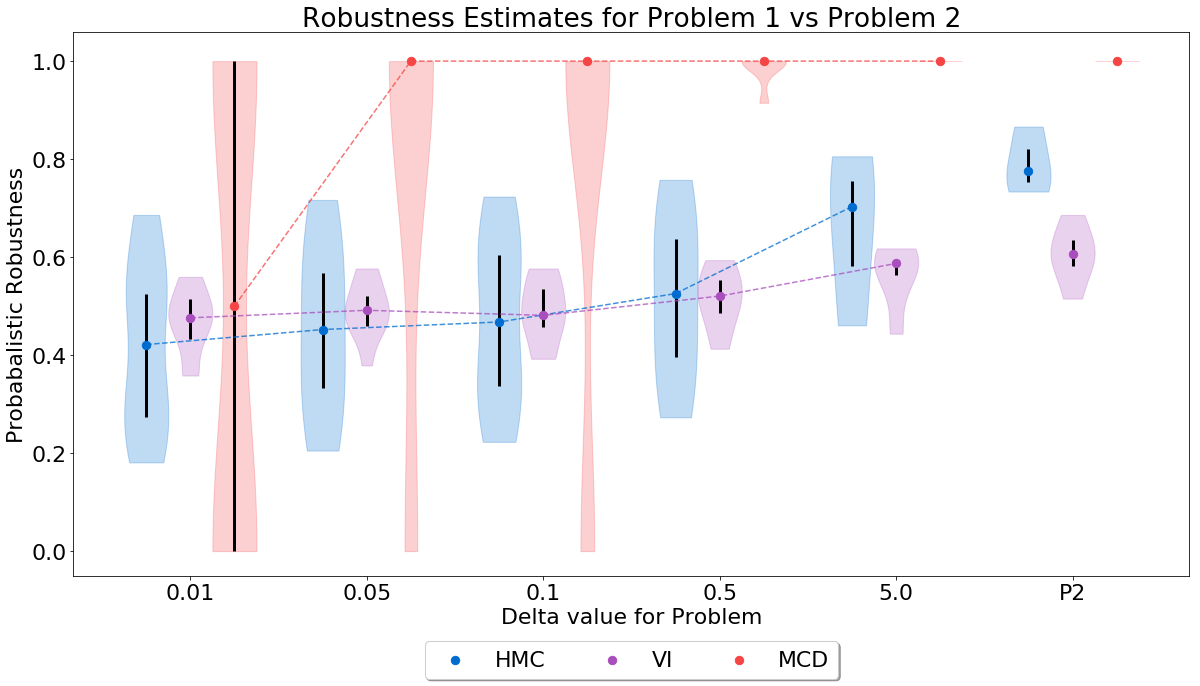}
\caption{In this figure we demonstrate the difference between Problems 1 and 2. To generate the statistics in this plot we extended the reachability analysis shown in Figure \ref{fig:verification-heatmaps}.}
\label{fig:Prob1vProb2}
\end{figure}

In Figure \ref{fig:Prob1vProb2}, we extend the tests on robustness to bounded perturbations that appear in Figure \ref{fig:verification-heatmaps}. It is clear to see that solving Problem 2 is, in some sense, similar to solving Problem 1 with an implicit value of $\delta$, where that value is approximately the value needed to yield a misclassification.

\begin{figure*}[h]
\centering
\includegraphics[width=0.95\textwidth]{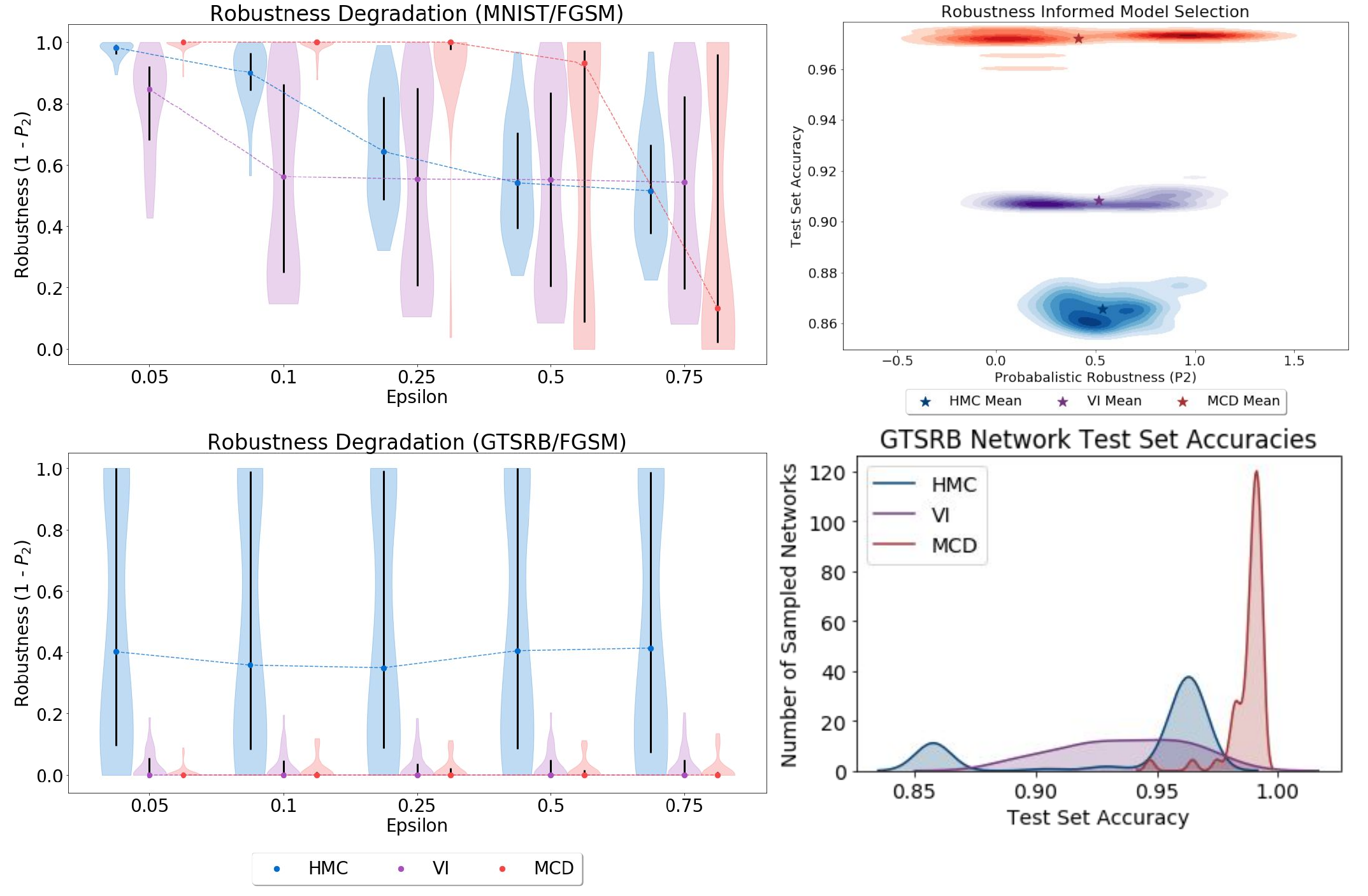}
\caption{We supplement the findings presented in figures 2 and \ref{fig:GTSRB-Attacks} by running the exact same experiment but using the gradient based attack proposed in \protect\cite{goodfellow6572explaining}. We note that though there is a slight quantitative variation between the figures (which is explained by the relative weakness of FGSM when compared to PGD), the qualitative behaviour in the robustness-accuracy trade-off is identical between the two tests.}
\label{fig:MNIST-Appendix}
\end{figure*}

\end{document}